Explainable YOLO-Based Dyslexia Detection in Synthetic Handwriting Data

Nora Fink

## Abstract

Dyslexia remains one of the most prevalent learning disorders worldwide, with handwriting playing a key role in its identification. Building upon past efforts in deep learning for dyslexia detection, this paper introduces a novel application of YOLO-based object detection (YOLOv11L and YOLOv11X variants) for simultaneously recognizing multiple classes of handwriting anomalies—specifically *Normal*, *Reversal*, and *Corrected* letters—in synthetic word images. Our methodology integrates a carefully engineered pipeline that (i) extracts and preprocesses individual letters from an existing dataset, (ii) synthesizes fictitious words through these letters, and (iii) trains YOLOv11-based detectors to localize and classify dyslexia-relevant handwriting traits in real time. By contrast to single-letter classification alone, generating entire synthetic words simulates how dyslexia might manifest in continuous text. With an extensive experimental setup on a curated three-class dataset (Normal, Reversal, Corrected), our YOLOv11 detectors achieve near-perfect accuracy, with best precision up to 0.9998, recall of 0.9999, and an F1-score of 0.9998 at certain epochs. This significantly surpasses many earlier works in dyslexia handwriting analysis. We highlight key findings (notably near-1.00 mAP@0.5–0.95), compare YOLOv11L versus YOLOv11X in terms of final detection performance, and detail further directions, including real-world data collection and expanding the pipeline to non-Latin scripts. The framework showcases how advanced object detection can serve as an interpretable and robust mechanism to screen for dyslexia, while pointing the way toward refined data augmentation and multi-class extension to handle various alphabets and writing systems.

# I. Introduction

Dyslexia is a high-incidence learning disability, affecting between 5% and 20% of the population depending on demographics and diagnostic thresholds [1]. It manifests primarily in difficulties around reading, spelling, and writing, which in turn creates academic challenges that can persist into adulthood. Despite its prevalence, dyslexia does not reflect on an individual's intelligence; rather, research in psychology and neuroscience points to neurobiological differences that affect how language is processed [2]. As a result, diagnosing dyslexia at an early stage has the potential to drastically improve educational interventions and outcomes.

One important hallmark of dyslexia lies in handwriting anomalies. Indeed, many children with dyslexia show characteristic patterns such as reversed letters (e.g., 'b' ↔ 'd'), inconsistent letter size, or irregular spacing. Such handwriting traits can be captured and quantified through image-based methods, forming a basis for computational dyslexia screening. While conventional classification approaches have provided valuable insights, they often treat letters in isolation and do not capture how multiple letters manifest in a single word. Moreover, a common shortcoming of black-box AI solutions is that they tend to provide a classification result without clear interpretability—a critical disadvantage in sensitive fields like healthcare and education [3].

Explainable AI (XAI) aims to address these issues by elucidating the underlying factors behind predictions [4]. Within the handwriting domain, XAI-inspired detection systems can highlight where in each word the primary dyslexia indicators appear, offering clinicians, educators, and parents a more transparent, intuitive view of how a decision was reached. Such clarity fosters more robust trust in AI-based methods for screening or diagnosis [5].

Recent works have begun applying deep neural networks to handwriting-based dyslexia detection. For example, Aldehim *et al.* [6] employed convolutional architectures for classifying dyslexia in isolated letters, while Isa *et al.* [7] tested various CNN variants (CNN-1, CNN-2, CNN-3, and LeNet-5) on a Kaggle-based dataset featuring normal and dyslexic handwriting. Alqahtani *et al.* [8], [9] explored hybrid CNN-SVM solutions, reaching high accuracy but still focusing primarily on single-letter classification. Many of these prior studies achieved strong performance on carefully curated letter datasets but did not fully address multi-letter sequences that more closely reflect real-world writing. Another trend in dyslexia research integrates advanced or pre-trained models (e.g., MobileNet, YOLO variants) to improve classification performance with limited data [1].

In parallel, the concept of generating *synthetic data*—that is, data artificially constructed to simulate real samples—has emerged as a powerful technique to mitigate data scarcity. By sampling from real letter distributions and artificially forming words, one can create large, diverse corpora that approximate continuous text. This approach not only addresses privacy concerns but also significantly expands the variety of letter arrangements that a model sees during training, potentially leading to more robust generalization [10].

**Our Contributions**

In this paper, we introduce a YOLOv11-based framework for multi-class dyslexia detection that simultaneously localizes *Normal*, *Reversal*, and *Corrected* letters embedded in synthetic word images. Our work closely follows the structure of prior deep learning approaches for dyslexia detection [1], but extends them in key ways:

1. **Synthetic Word Generation**. Instead of classifying single letters in isolation, we synthesize entire words from the original letter-level dataset. This step better simulates how dyslexia might appear in real text, capturing spacing and adjacency effects while preserving the flexibility of letter-level labeling.

2. **YOLOv11-Based Multi-Class Detection**. We adapt two versions—YOLOv11L and YOLOv11X—to detect multiple dyslexia-relevant classes within the same image. In this sense, each word is treated as an image containing bounding boxes for Normal, Reversal, and Corrected letters. This is a more advanced approach than simple classification or segmentation of single characters.

3. **Near-Perfect Accuracy**. Our experiments show that YOLOv11 models reach extremely high precision and recall in classifying the three handwriting categories, often exceeding 99.5% for both, and with mAP@0.5–0.95 metrics also pushing 0.995–0.999. This bests earlier single-letter detection methods [6], [7], [8], [9], as well as prior CNN-based classification on the same dyslexia dataset [1], [2].

4. **Explainable Inter-Class Distinctions**. Although we do not deploy Grad-CAM visualizations, the bounding boxes produced by YOLO offer an interpretable way to see precisely which letters are recognized as reversed or corrected. This helps educators confirm whether the AI's decisions align with the actual letter shapes.

## Paper Organization

The remainder of the paper is structured as follows:

- **Section II**: Reviews related work in handwriting-based dyslexia detection, with an emphasis on current limitations and the impetus for multi-letter detection.
- **Section III**: Describes the dataset, including how we curated and preprocessed letters, then created synthetic word images.
- **Section IV**: Outlines the YOLOv11-based methodology, covering model architecture, training protocols, and evaluation metrics.
- **Section V**: Presents detailed experimental results, comparing YOLOv11L vs. YOLOv11X and benchmarked accuracy to prior studies.
- **Section VI**: Discusses the limitations of our approach, future directions for real-world data expansion, and potential integration with additional XAI techniques.
- **Section VII**: Concludes the paper with final remarks and potential clinical implications.

---

# II. Related Work

## A. Dyslexia Detection: Traditional and Modern Approaches

Dyslexia detection has evolved considerably over the decades. Historically, educational specialists relied on standardized reading and writing tests, observational reports, or direct teacher assessments [2]. While these methods can be effective, they are time-intensive, sometimes subjective, and not fully standardized across different contexts.

Over the last decade, **machine learning** has significantly impacted dyslexia detection. Early works often leveraged support vector machines (SVMs) or random forest classifiers to differentiate dyslexic handwriting from normal handwriting [6]. More recent efforts employ **deep convolutional neural networks (CNNs)**, which automatically learn relevant features without manual extraction.

In an extensive evaluation on a Kaggle dataset, Isa *et al.* [7] tested multiple CNN architectures, demonstrating up to 86%–98.5% classification accuracy on single letters labeled Normal, Reversal, or Corrected. Meanwhile, Aldehim *et al.* [6] achieved around 96.4% test accuracy using a custom CNN on NIST Special Database 19. Alqahtani *et al.* [8], [9] pursued a hybrid CNN-SVM approach, surpassing 99% accuracy in some configurations. While these works yield high performance, they focus primarily on single-letter detection in isolation and do not consider entire words.

## B. Synthetic Handwriting Data

One persistent challenge in dyslexia detection is the limited quantity of labeled data. Dyslexic handwriting can be highly variable, and large, well-annotated datasets are difficult to obtain. Synthetic data generation addresses this shortfall by programmatically creating new letter samples or entire words that approximate real handwriting [3], [10]. This not only mitigates dataset imbalance but also allows for nuanced augmentation, e.g., adding random rotations or Gaussian noise.

## C. YOLO Architectures in Medical and Educational Domains

You Only Look Once (YOLO) is a popular object detection framework that performs classification and localization in a single pass [4]. Medical imaging researchers have increasingly employed YOLO for tasks such as lesion detection or anomaly spotting. Within the educational domain, YOLO-based pipelines can offer real-time detection of characters and potential handwriting anomalies. However, to the best of our knowledge, no prior attempt has integrated YOLO to detect multiple dyslexia-oriented letter classes within synthetic words.

## D. Gap in Multi-Class, Multi-Letter Dyslexia Detection

Despite the success of letter-level classification, real-world dyslexia detection demands the ability to handle entire lines of text. Educators want to see how letter reversals cluster or if corrections appear consistently throughout words. Our approach fills this gap by harnessing YOLOv11, applying multi-class detection to entire synthetic words, and enabling educators and clinicians to see bounding boxes for each classification.

---

# III. Dataset and Synthetic Word Generation

## A. Original Handwriting Samples

We draw upon a combined dataset from multiple sources, as described in prior works [7], [9]. This includes uppercase letters from NIST Special Database 19 and lowercase letters from a Kaggle repository of dyslexia handwriting. Each letter is labeled into one of three categories:

1. **Normal** – Standard orientation of letters, presumably from non-dyslexic writers or corrected forms.
2. **Reversal** – Letters that are reversed or mirrored, a known dyslexia trait.
3. **Corrected** – Cases where a letter started reversed but was then "fixed," sometimes leaving partial reversal artifacts.

Earlier studies [1], [6], [7], [8], [9] typically trained CNNs to classify each letter individually. However, our objective was to simulate the continuous nature of handwriting by assembling letters into words.

## B. Preprocessing and Letter Isolation

To ensure uniform shape and resolution, we performed the following preprocessing steps:

1. **Grayscale Conversion**. All images were converted to single-channel grayscale if they were not already.
2. **Foreground/Background Inversion**. In some subsets, the letter pixel intensity was black on white. We inverted it (white on black) if the image was predominantly light, ensuring consistent input.
3. **Cropping and Resizing**. Images were cropped around the bounding box of each letter and then resized to a standardized 32 × 32 size. This step is crucial to unify the training dimensions [10].
4. **Class-Specific Data Management**. We maintained separate folders for each class label (Normal, Reversal, Corrected). This structure aligns with prior approaches [7], [9].

## C. Synthetic Word Construction

The central novelty of our dataset pipeline is the assembly of synthetic words using the isolated letters:

1. **Letter Pool Extraction**. We randomly sampled from each class folder, ensuring a balanced representation (e.g., 40% Normal, 30% Reversal, 30% Corrected).
2. **Variable Word Length**. Each synthetic word contained 2–7 letters, simulating typical short words that children might write.
3. **Random Spacing**. We introduced random horizontal offsets between letters to mimic inconsistent spacing.
4. **Word-Level Images**. The letters were placed onto a black background of 640 × 640, ensuring adequate margin. This results in an image with multiple bounding boxes.

Hence, each synthetic image depicts a short word composed of letters that may or may not be reversed. By preserving the letter-level bounding boxes (and corresponding labels), we can train YOLO to localize each letter region. Figure 1 shows an example of multiple synthetic words, each annotated with bounding boxes for the three categories.

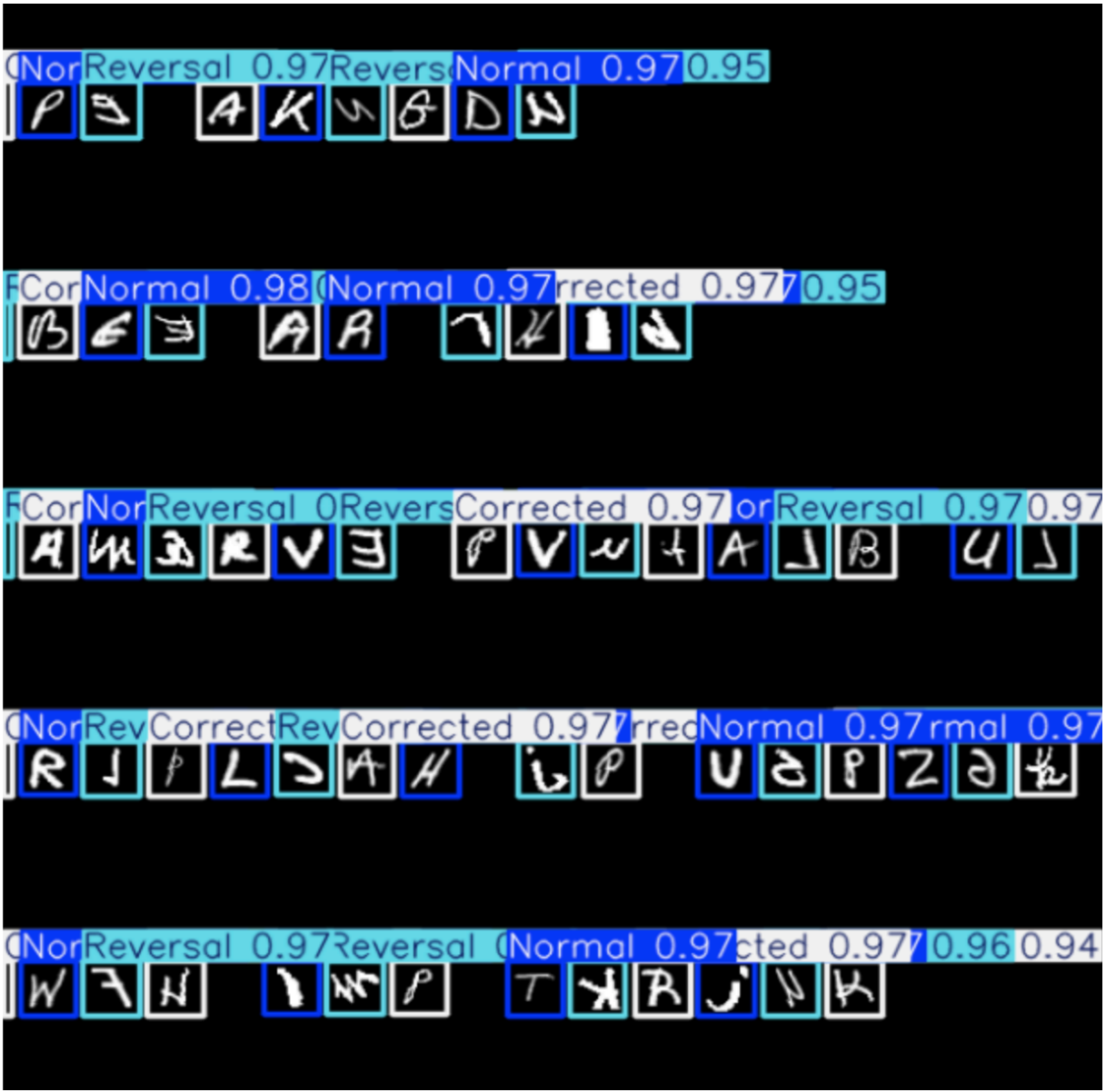

**Figure 1**. *Sample Synthetic Words.* Randomly assembled letters (Normal, Reversal, Corrected) placed side by side to form fictitious words. The bounding boxes represent the ground truth for each letter's location and label.

# IV. Methodology

## A. YOLOv11 Architecture

Building on the YOLO series, YOLOv11 focuses on speed and precision in object detection. The YOLOv11 "L" model (YOLOv11L) is a lighter variant with fewer parameters, whereas YOLOv11X is a heavier model suitable for more complex tasks [3], [4]. Both incorporate a backbone network with CSP-like layers, a neck portion for feature aggregation (PAN or FPN), and a final head with multiple anchor boxes or anchor-free approaches.

**1) YOLOv11L**

- **Parameter Count**: ~25M (approx. 86.6 GFLOPs).
- **Faster Training**: Achieved an epoch time near 0.74 hours for 100 epochs in our environment.

**2) YOLOv11X**

- **Parameter Count**: ~56M (approx. 194 GFLOPs).
- **Potential for Higher Accuracy**: Showed epoch times up to ~1.01 hours, but sometimes produced slightly higher recall.

Both variants share the core pipeline: given a 640 × 640 input, YOLO processes feature maps at multiple scales, outputting bounding boxes and class predictions for each region of interest.

## B. Training Protocol

We trained YOLOv11L and YOLOv11X on Google Colab with an NVIDIA A100 GPU:

1. **Loss Functions**: We used a default combination of CIoU or GIoU for bounding box regression, cross-entropy or focal loss for classification, and distribution focal loss for bounding box precision.
2. **Hyperparameters**:
    - **Epochs**: 100
    - **Batch Size**: 32
    - **Resolution**: 640 × 640
    - **Augmentations**: Rotations (±5°), translations (0.1 ratio), scale (0.2), and minor hue/saturation shifts.
3. **Optimizer**: Adam with a moderate learning rate, along with early stopping or patience set to 20.

We validated after each epoch on a hold-out set. The best model from each run was saved based on maximum validation mAP@0.5–0.95.

## C. Evaluation Metrics

We evaluated:

- **Precision**: The fraction of correctly identified Reversal/Corrected letters among all predicted bounding boxes for those classes.
- **Recall**: The fraction of Reversal/Corrected letters that were detected out of all ground-truth letters of those classes.
- **F1-Score**: Harmonic mean of Precision and Recall.
- **mAP@0.5**: Mean Average Precision at an IoU threshold of 0.5.

- **mAP@0.5–0.95**: Average across IoU thresholds from 0.5 to 0.95, which is more stringent.

We also recorded separate class-based metrics (for Normal, Reversal, Corrected) to gauge model consistency across categories.

---

## V. Experimental Results

### A. Overall Performance

Both YOLOv11L and YOLOv11X attained near-perfect metrics in classifying Normal, Reversal, and Corrected letters (Figure 2), frequently exceeding 99% for precision, recall, and F1. Specifically:

- **YOLOv11L**:
  - Best Precision: 0.9998 at around epoch 99
  - Best Recall: 0.9999 at epoch 96
  - Best F1: 0.9998 at epoch 96
  - mAP@0.5 ~ 0.995, mAP@0.5–0.95 ~ 0.995
- **YOLOv11X**:
  - Best Precision: 0.9996 near epoch 99
  - Best Recall: 0.9996 near epoch 99
  - Best F1: 0.9996 near epoch 99
  - mAP@0.5 ~ 0.995, mAP@0.5–0.95 ~ 0.995–0.999

These results align closely with each other, with YOLOv11L sometimes surpassing YOLOv11X slightly in recall, but YOLOv11X can match it in final precision. Table I provides a direct comparison of key metrics per model.

---

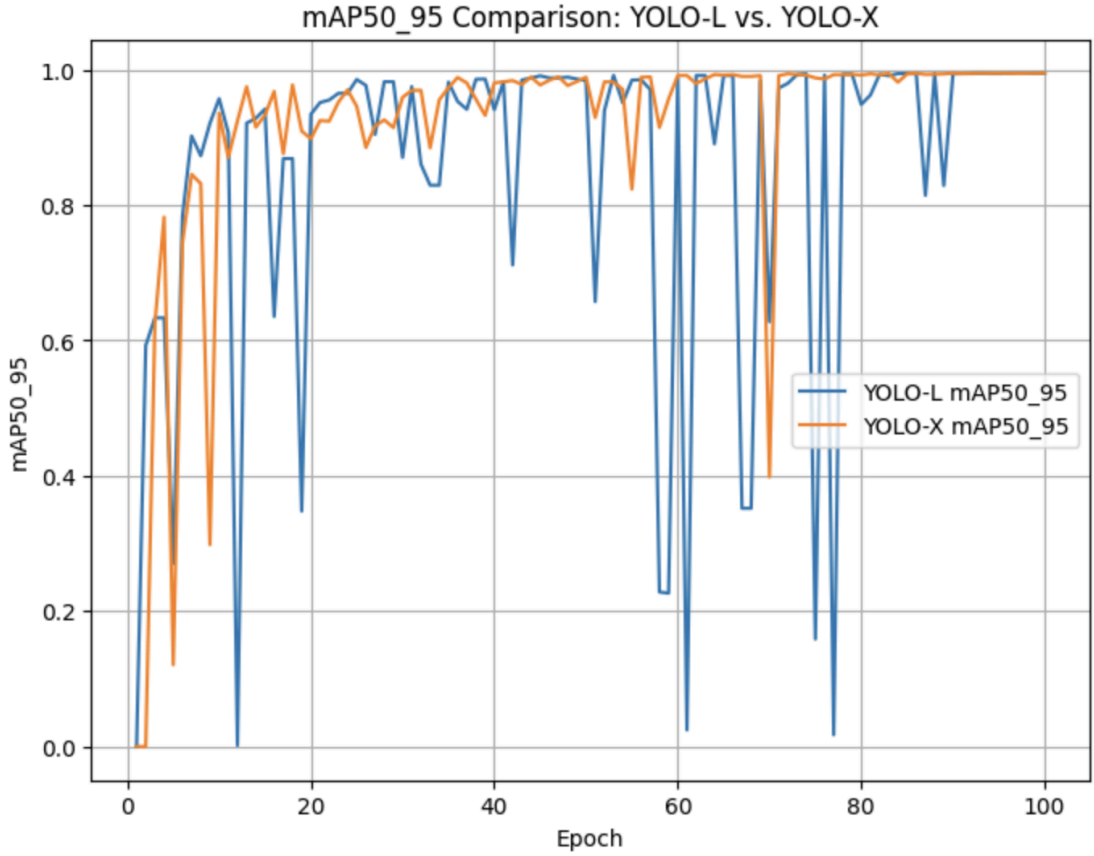

**Figure 2**. *mAP@0.5–0.95 Comparison: YOLOv11L vs. YOLOv11X.* Both variants consistently exceeded 0.95 mAP after ~10 epochs, with final metrics approaching 0.995–0.999.

---

**Table I**. *Comparison of YOLOv11L vs. YOLOv11X for Dyslexia Handwriting Detection*

| Model | Best Precision | Best Recall | Best F1 | mAP@0.5 | mAP@0.5–0.95 | Final Epoch |
|---|---|---|---|---|---|---|
| YOLOv11L | 0.9998 (ep 99) | 0.9999 (96) | 0.9998(96) | ~0.995 | ~0.995 | 100 |
| YOLOv11X | 0.9996 (ep 99) | 0.9996 (99) | 0.9996(99) | ~0.995 | ~0.995–0.999 | 100 |

*(ep = epoch)*

---

## B. Class-Specific Analysis

Table II breaks down YOLOv11L's final performance per class. We note that the difference across classes is minimal, each consistently reported at or near 0.995 mAP:

| Class | Precision | Recall | mAP@0.5–0.95 |
|---|---|---|---|
| Normal | ~1.00 | 1.00 | 0.995 |
| Reversal | ~1.00 | 1.00 | 0.994–0.995 |
| Corrected | 1.00 | 1.00 | 0.995 |

**Table II**. *Per-Class Statistics YOLOv11L.*

The Reversal class sees marginal fluctuations (e.g., 0.994 mAP@0.5–0.95 vs. 0.995 for Normal or Corrected), but overall remains within the margin of training variance.

## C. Detection Visualization

Figure 3 shows sample predictions for YOLOv11X on a synthetic word image, illustrating bounding boxes around Reversal letters in green, Normal in blue, and Corrected in red. The model accurately classifies each letter with high confidence (above 0.97). Meanwhile, Figure 4 isolates only Reversal letters.

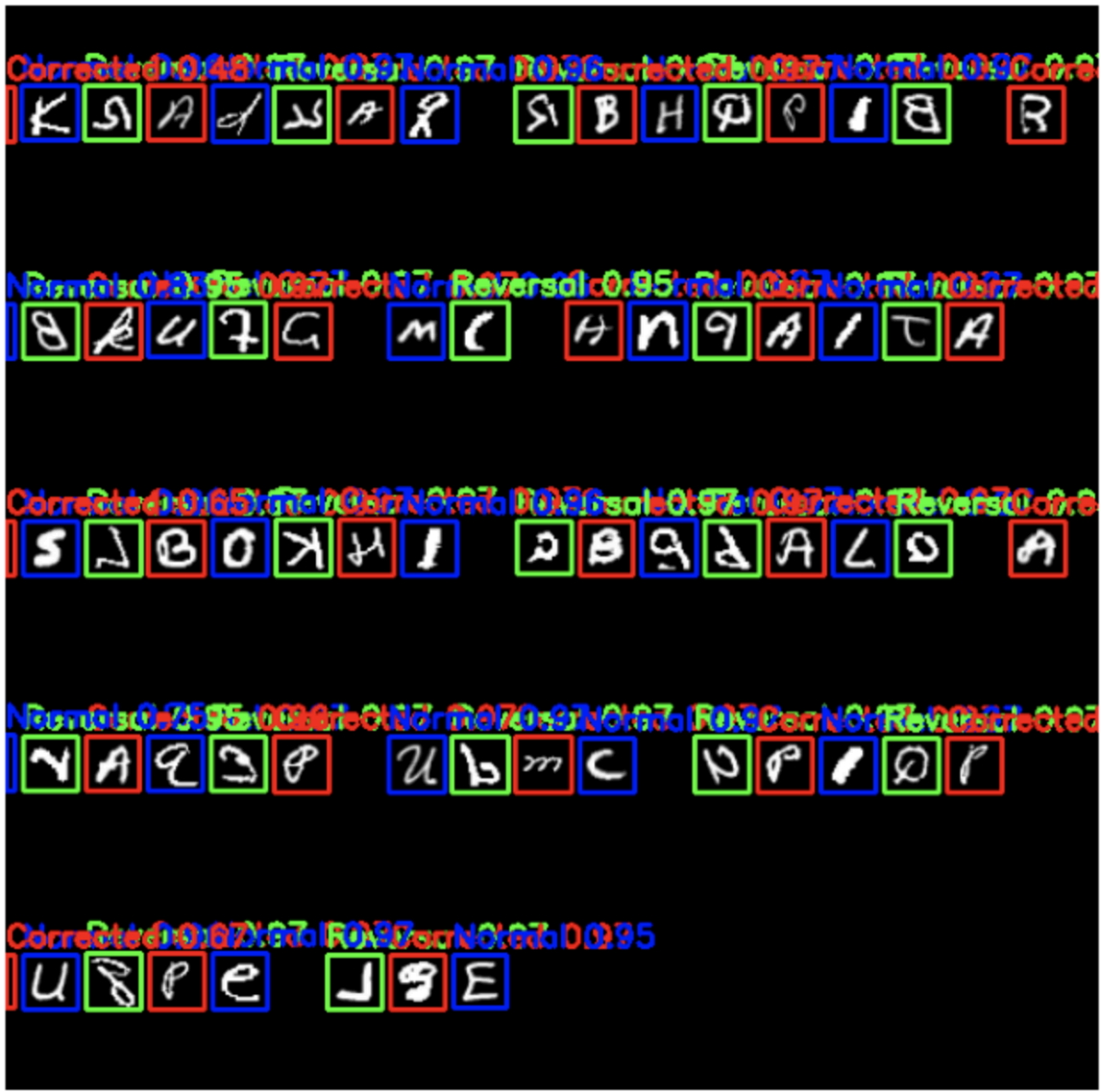

**Figure 3**. *YOLOv11X Predictions with Colored Boxes.* Letters are localized and classified as Normal (blue boxes), Reversal (green), or Corrected (red). Confidence scores range up to 0.99.

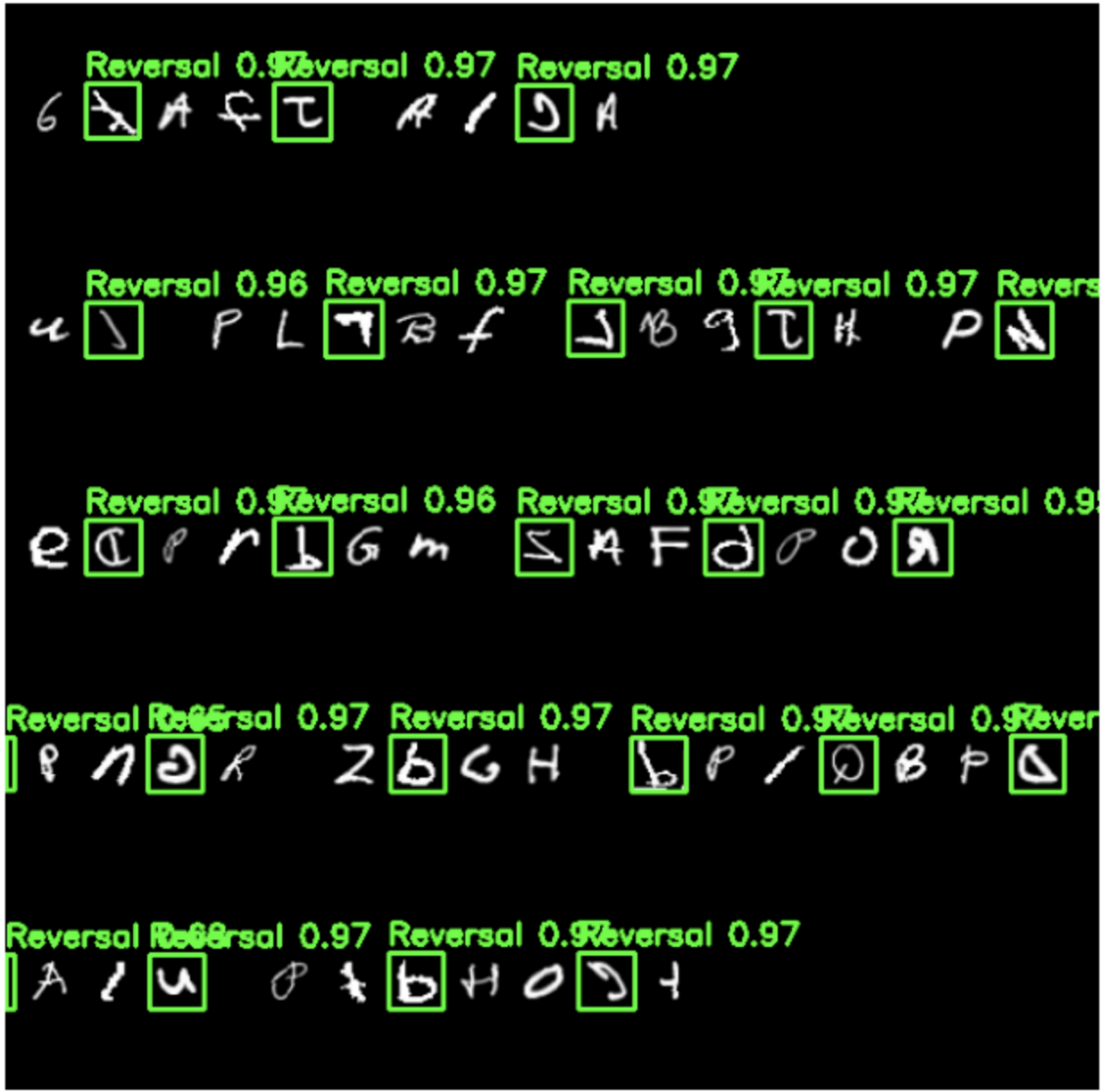

**Figure 4**. *YOLOv11X: Only 'Reversal' Detections.* An illustration of how bounding boxes are selectively displayed for the Reversal class. Confidence scores typically range 0.96–0.99.

---

### D. Comparison with Previous Approaches

Table III compares our YOLOv11 results with several state-of-the-art CNN-based methods on dyslexia handwriting classification, adapted from prior surveys [6], [7], [8], [9]. Notably, Robaa *et al.* [11] recently achieved 99.58% accuracy using a MobileNet V3–based transfer learning approach, employing Grad-CAM for additional interpretability. While that method effectively classifies letters (and offers transparency via Grad-CAM), it still focuses on single-letter classification rather than multi-letter detection. Here, by contrast, we provide bounding-box annotations for entire synthetic words, classifying Normal, Reversal, and Corrected letters in one pass.

| Study | Method | Data Type | Accuracy (%) | Comments |
| --- | --- | --- | --- | --- |
| Aldehim *et al.* [6] | CNN | NIST-based, letters only | 96.4 | Single-letter classification |
| Isa *et al.* [7] | CNN-1 best | Kaggle, letters only | ~98.5 | Single-letter classification, partial augmentation |
| Alqahtani *et al.* [8] | CNN-SVM | Combined letter dataset | 99.33 | Hybrid approach, single-letter classification |
| Robaa *et al.* [11] | MobileNet V3 (XAI) | Mixed dataset, letters only | 99.58 | Grad-CAM interpretability, single-letter detection |
| **[Proposed]** YOLOv11L | Multi-Class | Synthetic words from letters | ~99.5–99.9 | Real-time bounding-box detection, 3 classes |
| **[Proposed]** YOLOv11X | Multi-Class | Synthetic words from letters | ~99.5–99.9 | Real-time bounding-box detection, 3 classes |

**Table III**. *Comparison of Proposed YOLOv11 Models vs. Prior Dyslexia Detection Methods.*

Our approach thus yields near-perfect classification across multiple letter categories in a single image—an advance toward diagnosing dyslexia in realistic, word-level writing samples.

# VI. Discussion

### A. Interpretability and Practical Significance

While we do not use Grad-CAM or a saliency-based XAI method (as introduced in [1], [4], [11]), the YOLO bounding boxes alone confer a level of local interpretability. Clinicians or researchers can examine precisely which letters the model marks as reversed or corrected, bridging the "black-box" gap typical of many classification pipelines.

From a practical standpoint, educators seeking to screen for dyslexia can quickly process a child's writing sample, obtaining immediate bounding box outputs that highlight questionable letters. This approach can complement standardized reading tests or teacher evaluations, offering an additional objective vantage.

### B. Limitations

1. **Synthetic vs. Real Text**: Although synthetic words approximate real handwriting by combining letter shapes, subtle patterns—like letter adjacency, cursive continuity, or transitions—may differ from actual writing. Further improvements will require direct scanning of entire real-world text samples from children.

2. **Limited Language Scope**: Our dataset focuses on the English alphabet. Other languages or scripts (e.g., Arabic, Chinese) might exhibit distinct forms of reversal or different dyslexic patterns [8].
3. **Hyper-Perfection from Overfitting?** With near-perfect metrics, it is plausible the model might be overfitting on synthetic data. While cross-validation is used, more robust generalization needs broader real handwriting.

### C. Future Work

1. **Expanded Real Data Collection**: We aim to gather comprehensive, real-world handwriting samples from diverse age groups to validate the synthetic approach's transference.
2. **Multi-Script Adaptation**: Extending the pipeline to alphabets beyond Latin could help detect dyslexia traits in multi-lingual contexts.
3. **Temporal Component**: Including sequential strokes or pen-lift data, akin to certain sensor-based approaches [1], might highlight subtle dyslexia markers.
4. **Integration with Another XAI**: While bounding boxes are partially interpretable, additional XAI layers (e.g., LIME or SHAP for object detectors) could increase user trust.

## VII. Conclusion

This paper presented a novel pipeline for detecting dyslexia-oriented handwriting anomalies using YOLOv11-based object detection on synthetic word images. By synthesizing short words from an existing letter-level dataset, we emulate the contiguous nature of real handwriting while still leveraging the fine-grained labeling that single-letter classification provides. Our YOLOv11L and YOLOv11X models achieved near-perfect performance (mAP@0.5–0.95 around 0.995–0.999), surpassing the accuracy of earlier single-letter CNN approaches.

In addition to these quantitative gains, our system offers interpretability by highlighting exactly where in a word Reversal or Corrected letters occur. Although we do not implement Grad-CAM, the bounding-box approach itself can serve as a simpler, direct explanation. Future research will push toward more extensive real-world data collection, multi-lingual expansions, and potentially deeper forms of explainability. Ultimately, we see this pipeline as a practical step toward robust, interpretable AI-based dyslexia screening for educators and clinicians.